\documentclass[journal]{IEEEtran}

\usepackage{amsmath,amssymb,amsfonts}
\usepackage{algorithmic}
\usepackage{graphicx}
\usepackage{textcomp}
\usepackage{xcolor}
\usepackage{stfloats}
\usepackage{array}
\usepackage{multirow}
\usepackage{algorithm}  
\usepackage{algorithmic} 
\usepackage{subcaption}

\ifCLASSINFOpdf

\else

\fi

\begin{document}

\title{Causality-Based Reinforcement Learning Method for Multi-Stage Robotic Tasks}

\author{Jiechao Deng, and~Ning Tan*,~\IEEEmembership{Member,~IEEE},
\thanks{J.C. Deng and N. Tan are with the School of Computer Science and Engineering, Sun Yat-sen University.}
\thanks{This work was supported in part by the National Natural Science Foundation of China under Grant 62173352, and in part by the Guangdong Basic and Applied Basic Research Foundation under Grant 2024B1515020104.}
}

\maketitle

\begin{abstract}
Deep reinforcement learning has made significant strides in various robotic tasks. However, employing deep reinforcement learning methods to tackle multi-stage tasks still a challenge. Reinforcement learning algorithms often encounter issues such as redundant exploration, getting stuck in dead ends, and progress reversal in multi-stage tasks. To address this, we propose a method that integrates causal relationships with reinforcement learning for multi-stage tasks. Our approach enables robots to automatically discover the causal relationships between their actions and the rewards of the tasks and constructs the action space using only causal actions, thereby reducing redundant exploration and progress reversal. By integrating correct causal relationships using the causal policy gradient method into the learning process, our approach can enhance the performance of reinforcement learning algorithms in multi-stage robotic tasks.

\end{abstract}


\IEEEpeerreviewmaketitle

\section{Introduction} \label{section1}

\begin{figure}[]
	\centerline{\includegraphics[scale=0.6]{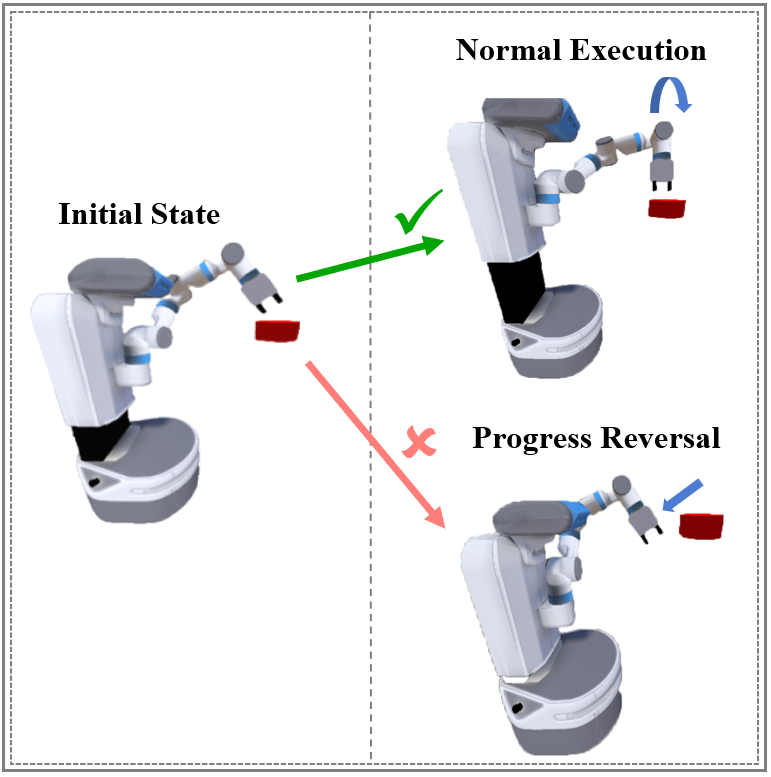}}
	\caption{During normal execution, the end effector should be adjusted to face downwards, but due to the movement of the robotic arm, the position of the end effector deviated, causing the entire task to regress from step 2 back to step 1, resulting in a progress reversal.}
	\label{progress revesal}
\end{figure}

\IEEEPARstart{D}{eep} reinforcement learning (RL) has achieved many results in robot tasks in recent years, such as local obstacle avoidance \cite{duguleana2016neural}, indoor navigation \cite{kulhanek2019vision}, door opening \cite{gu2017deep}, object manipulation \cite{andrychowicz2017hindsight}, and mobile manipulation \cite{wang2020learning}.
However, in robot related tasks, there are many tasks that need to be completed in stages. When faced with such multi-stage tasks, it is difficult to simply use a single agent for RL to solve problems. In multi-stage tasks, agents need to learn both the immediate results of their actions and how these results affect the completion of the entire task, and these contents are often intertwined. When dealing with long-horizon tasks that require multiple stages to complete, RL algorithms often explore into dead ends, and may even lead to a reversal of progress.\cite{hundt2020good}. Progress reversal refers to the situation where a task is divided into multiple stages, the actions of subsequent stage may alter the state achieved by the previous stage, leading to the incompleteness of the previous stage.

As shown in Fig. \ref{progress revesal}, the robot's end effector has already moved above the target point. The next step should be to rotate the end effector downward. However, if the end effector also moves during the rotation, this will result in a reversal of progress, requiring the end effector to be moved back above the target point. If the aforementioned robot has the capability to both move and rotate its end effector, and if it does not perform the movement operation while rotating, it can significantly avoid the occurrence of progress reversal. In real-life scenarios, humans, when executing tasks, instinctively select actions that they know will influence the task's completion, based on their comprehension of how their actions affect the surrounding environment. Upon observing the outcomes of these actions, they make fine-tuned adjustments to their subsequent movements. That is, if one can obtain the causal relationship between actions and the environment, it is possible to execute only the necessary actions to complete the task, and adjust oneself based on the feedback results corresponding to the actions.

In RL, reward variables can be used as indicators of task completion. By identifying the causal relationship between actions and rewards, one can select the causal actions for each stage based on the rewards of that stage to construct the action space of the deep RL policy. This approach can reduce redundant exploration and the occurrence of progress reversal. Furthermore, causal policy gradient method \cite{hu2023causal} can utilize known causal relationships to allow policy to learn based on the rewards corresponding to actions, thereby reducing the gradient variance in the RL process. Therefore, if accurate causal relationships for each stage of multi-stage tasks can be obtained, the performance of RL in handling multi-stage tasks can be enhanced by utilizing causal actions and causal policy gradients.

Based on this, we propose a method that leverages causal relationships and deep RL to handle multi-stage tasks. This method allows the robot to autonomously discover the causal effects of its actions on task completion and constructs the action space of the deep RL algorithm using only causal actions. By integrating causal policy gradients for learning, it can effectively handle multi-stage tasks.
The contributions of this study can be summarized as follow:
\begin{enumerate}
 \item Approach multi-stage tasks from the perspective of breaking down subtasks. By leveraging the robots' ability to interact with the environment to establish causal relationships for each subtask, thereby obtaining the causal relationships between the robots' actions and the environment at each stage.
  \item A RL method for multi-stage tasks was designed using the obtained causal relationships.This method can reduce redundant exploration in RL by narrowing the action space, decrease the instances of progress reversal in multi-stage tasks, and enhance the performance of RL.
  \item Conduct experimental comparisons of the proposed method on a mobile manipulation task and a pure manipulation task to verify the effectiveness of the proposed approach.
\end{enumerate}

\section{RELATED WORK} \label{section2}
Various approaches have been proposed for multi-stage tasks, some of these methods still employ a single agent for learning multi-stage tasks.
The direct physical consequences of multi-stage task behavior are entangled with the impact of these consequences on the overall goal, making robot task learning difficult. In the process, RL wastes significant time exploring unproductive actions, such as spending a lot of time grabbing air in tasks like stacking blocks.
\cite{hundt2020good} proposed the SPOT framework for more efficient RL of multi-stage tasks.
\cite{stengel2022guiding} propose a novel Transformer-based model which enables a user to guide a robot arm through a 3D multi-step manipulation task with natural language commands base on SPOT.
To efficiently address sequential object manipulation tasks,
\cite{bao2022learn} divide actions into different groups, assuming that some actions are preconditions of others in multi-step tasks, and apply pixel-wise Q value-based critic networks to solve multi-step sorting tasks.

The approach of decomposing tasks into multiple subtasks is inherent in hierarchical reinforcement learning (HRL), as its advantage lies in the ability to learn a series of simple and smaller subtasks, and then apply these subtasks to solve larger and more complex problems.
Using HRL, \cite{oh2017zero} enables the agent to learn and perform subtasks based on given instructions.
If changing the state that the robot needs to achieve in completing the task is regarded as each subtask, \cite{nachum2018data,li2020hrl4in} utilize HRL, enabling the upper-level meta-controller to set subtasks for the underlying policies to complete.

There are also many methods that split tasks into subtasks but do not utilize HRL.
With Hindsight Experience Replay (HER) \cite{andrychowicz2017hindsight}, the agent can generate non-negative rewards by goal relabeling strategy to alleviate the negative sparse reward problem.

\cite{luo2022relay} propose a novel self-guided continual RL framework, Relay-HER (RHER), utilizing HER and a Self-Guided Exploration Strategy(SGES) to solve sequential object manipulation tasks efficiently. 
\cite{erskine2022developing} also divides multi-stage tasks into multiple subtasks for learning. However, it proposes a strategy of not only training each subtask separately but also considering the situation of the latter subtask during training. This allows agents of multiple tasks to learn to cooperate with each other.
\cite{wang2023multi} introduce MRLM for non-prehensile manipulation of objects. MRLM divides the task into multiple stages according to the switching of object poses and contact points. It then employs an off-policy RL algorithm to train a point cloud motion based manipulation network (P2ManNet) as the policy to complete each stage successively.
\cite{andreas2017modular} addresses multi-task RL guided by abstract sketches of high-level behavior.

RL techniques based on causal modeling can leverage structural causal knowledge to enhance the performance of RL algorithms.
\cite{gasse2021causal} employs the do-calcules \cite{pearl2012calculus} to formalise model-based RL as a causal inference problem, and present a method that combining offline and online data in model-based RL. \cite{hu2022causality} utilizes causal discovery algorithms to identify causal relationships between actions and state variables. It then constructs multi-layered policies based on these causal relationships to facilitate HRL. \cite{hu2023causal} employs causality to address the mobile manipulation problem of robots. It use causal discovery algorithms to obtain the causal relationship between actions and rewards and applies the causal policy gradient method to update the policy.

We also approach the multi-stage robotic tasks by decomposing them into multiple subtasks and leveraging causal relationships.

\section{PROBLEM STATEMENT}
A N-stages task $T$ is modeled from the perspective of multi-reward MDP (\textbf{s}, \textbf{a}, p, \textbf{r}), where $\textbf{s}$ is the state space, $\textbf{a} = (a_1,\ldots,a_K)$ is the action space, $p$ is the markovian transition model, $\textbf{r}$ is a vector containing all reward items, i.e. $\textbf{r}=(r_1,...., r_M)$. We decompose each stage of the multi-stage task into a subtask, so $T$ is decomposed into N subtasks $\{ST_1, \ldots, ST_N\}$. Each subtask corresponds to a part of the reward term $\textbf{r}^i = (r_1^i,\ldots,r_J^i) \subseteq \textbf{r}, i \in [1,N]$. For subtask $ST_i, i \in [1,N]$, there is an agent $A_i$ consist of a policy $\pi_i$. Starting from the step the task enters stage $i$, the sequence of interactions between the agent and the environment through a series of actions until the task transitions to another stage is considered an episode of subtask $i$. Let $\tau_i$ represent all the time steps in this episode, $\textbf{r}^i_{t}$ represents the reward obtained for subtask $i$ at time step $t,t \in \tau_i$. Each $\pi _i$ needs to maximize its corresponding return $R_{i} =\sum_{t \in \tau_i} \gamma\textbf{r}^i_{t}$, $\gamma$ is the discount factor. At step $t$, $\textbf{s}_t$ represents the current environmental state, and $U(\textbf{s}_t) \in [1,N]$ is used to determine the current subtask. Invoke $A_{U(\textbf{s}_t)}$ based on the current subtask index $U(\textbf{s}_t)$, input $\textbf{s}_t$ into $\pi_{U(\textbf{s}_t)}$, output the action value to be executed, and obtain the reward $\textbf{r}^{U(\textbf{s}_t)}$. 

\begin{figure}[t]
  \centering
    \includegraphics[scale=0.5]{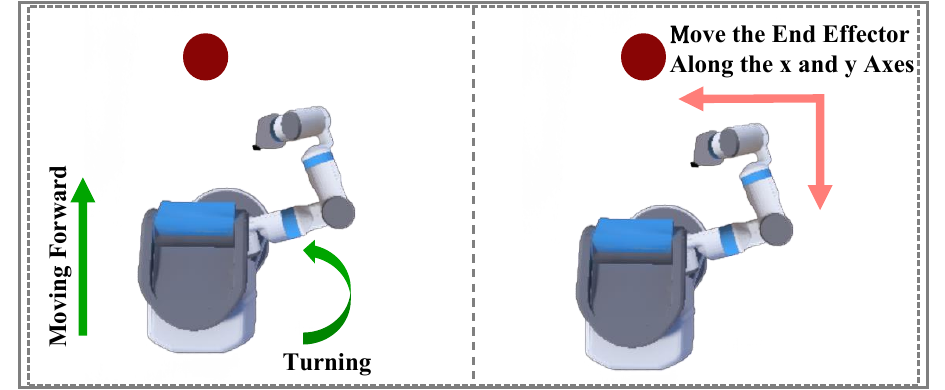}
   \caption{Two Action Selection Schemes}
  \label{Redundant action}
\end{figure}

In certain tasks, robots require only a subset of the available actions. For instance, as depicted in Fig.\ref{Redundant action}, to move the end effector to the red dot, the robot has five actions at its disposal: moving forward, turning, and moving along the x, y, and z axes. To achieve horizontal movement, one can either utilize just moving forward and turning, or solely the x and y axis movements. For each $\pi_i$, our goal is to select a set of actions that are both complete enough to accomplish the task in stage i and non-redundant to construct the action space to reduce redundant exploration and progress reversal in RL. Additionally, each action dimension should be updated based on the reward items it affects. Both objectives require the use of the causal relationship between actions and rewards.

\subsection{Causal Relationship}\label{Causal Relationship}
Structural causal modeling is a method that can formally express the causal assumptions behind the data and can be used to describe the correlation characteristics and their interactions in the real world\cite{pearl2016causal}. The core idea of Structural Causal Models is to describe the causal relationships between variables using Directed Acyclic Graphs. Each node represents a variable, and the directed edges indicate causal relationships, meaning that the arrow points from the cause to the effect.

The structural causal model contains two variable sets $U$ and $V$, and a set of functions
\begin{equation}
 f=\{f_x:W_x \rightarrow X|X \in V\}
 \end{equation}
where $W_x \subseteq (U \cup V)-\{X\}$. 
That is, function $f_x$ assigns a value to variable X based on the values of other variables in the model. Therefore, the definition of cause and effect is: if $Y$ is in the domain of $f_x$, then $Y$ is the cause of $X$. If node $X$ in the graph is a child node of another node $Y$, then $Y$ is the cause of $X$. The adjacency matrix used to represent this causal relationship in a directed acyclic graph is referred to as the causal matrix. Here, $U$ represents exogenous variables, which are determined by factors outside the model and are not influenced by other variables within the model. Our study focuses on the causal relationships among endogenous variables $V$. 

\subsection{Causal Action Space}\label{Causal Action Space}
To reduce redundant exploration in each subtask's agent during RL, one can seek out a complete yet non-redundant set of actions that are necessary to accomplish each subtask. The number of reward terms for each subtask can typically characterize the completion status of the agent's task. Therefore, finding actions that affect task completion is equivalent to finding actions that affect the task's reward terms. To this end, it is necessary to study the causal relationships between actions and reward terms. This study focuses solely on the causal relationships between action terms and reward terms, considering action variables as the causal variables and reward variables as the outcome variables. For subtask $ST_i$, $r_j^i=f_r(a_k)$ illustrating that $a_k$ is the causal of $r_j^i,j \in (1,\ldots,J)$. Causal Action $\textbf{ca}_i$ is:
\begin{equation}\label{eq:CausalActions}
 \textbf{ca}_i =\{a_k|(a_k \in \textbf{a}) \land (\exists r_j^i \in \textbf{r}^i,r_j^i=f_r(a_k))\}
 \end{equation}

Here, our focus lies in establishing the causal graph model rather than specifying the mapping function $f_r$. Therefore, our focus is on obtaining the causal matrix $m_i$
for $ST_i$. Using actions as the rows of the causal matrix and reward items as the columns, the causal actions for the subtask $ST_i$ are the actions corresponding to the rows where the sum of the column for $\textbf{r}^i$ is not zero. In the field of robot tasks, continuous actions are more common, so we discuss the scenario where the action space consists of continuous values.

\subsection{Causal Policy Gradient}\label{Causal Policy Gradient}
For each subtask $ST_i$, when obtaining causal actions, the causal matrix $m_i$ is first derived. $m_i$ can be integrated into RL using causal policy gradients to reduce the gradient variance during the learning process. In robotic tasks, the reward function is typically a composite form, and the reward value is the linear summation of reward terms corresponding to each goal.
However, by employing the causal policy gradient\cite{hu2023causal} method, different reward combinations for each action can be used. It integrating causality into the implementation of policy learning to reduce gradient variance. For $ST_i$, its corresponding agent $A_i$'s policy $\pi_i$ has parameters $\theta_i$, this approach redefines the policy gradient to be:
\begin{equation}
 \nabla_{\theta_i} J(\theta_i) = \nabla_{\theta_i} log\pi_{\theta_i} (ca_i|s)\cdot m_i \cdot \widehat{\mathbb{A}}^{\pi_i}(s,ca_i)
\end{equation}
$\widehat{\mathbb{A}}^{\pi_i}(s,ca_i)$ is the advantage function factored across the reward terms.

\section{METHOD} \label{METHOD}
\begin{figure*}[htbp]
	\centerline{\includegraphics[scale=0.50]{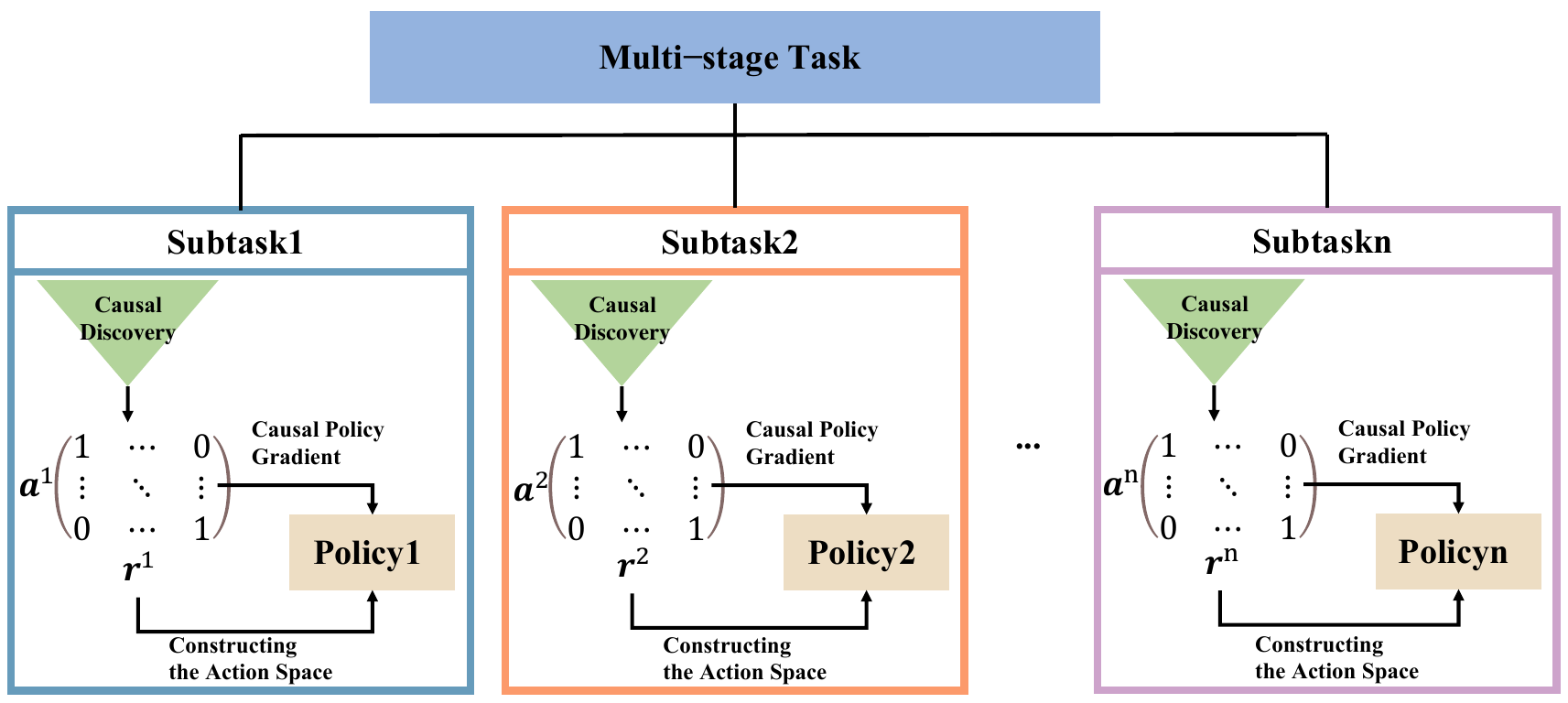}}
	\caption{Breaking down multi-stage tasks into multiple subtasks, each subtask discovers the causal relationships of the current stage, which are used to construct the action space of the current policy and are integrated into the learning process using the causal policy gradient method.}
	\label{task decompose}
\end{figure*}

Given a multi-stage task $T$, the agent has available actions $\textbf{a}$. Divide the reward  terms $\textbf{r}$ into $\{\textbf{r}^i|i \in \{1,\ldots,n\}\}$ according to different task stages. Each $\textbf{r}^i=\{r^i_1, \ldots, r_J^i\}$ is composed of one or more reward terms. For each subtask $ST_i,i \in \{1,\ldots, n\}$, the first step is to obtain its causal matrix $m_i$ between actions $\textbf{a}$ and reward terms $\textbf{r}^i$. Then, we use $m_i$ to construct the action space of the RL policy and integrate it into the policy learning using the causal policy gradient method. Therefore, our approach mainly consists of two major steps: automated causal matrix discovery and training. The framework of the method is shown in the Fig.\ref{task decompose}

\subsection{Causal Matrix Discovery}\label{Causal Matrix Discovery}
The causal models involved in this study only include two types of variables: action variables and reward item variables. Action variables serve only as cause variables, while reward item variables serve only as effect variables. Leveraging the characteristic of robots to interact with the environment while executing actions, we propose a method for obtaining causal relationships between actions and rewards, which can achieve more accurate causal relationships than those specified by humans. 

For subtask $ST_i$, the objective is to determine whether there is a causal relationship between $a_k,k \in \{1, \ldots, K\}$ and $r^i_j,j \in \{1,\ldots,J\}$, that is, to determine whether there exists a directed edge starting from vertex $a_k$ and ending at vertex $r^i_j$. 
This can be determined by having the robot execute different values of action $a_k$ and comparing the changes in the reward distribution under different action values.
Therefore, we need to intervene on $a_k$ to observe whether the intervention affects the distribution of $r^i_j$. Intervention refers to forcing a variable 
$X$ to take a specific fixed value $x$ \cite{pearl2009causality}, using $do(X=x)$ to denote setting the value of $X$ to $x$. In our design, the intervention sets a single action variable $a_k$ to a fixed value. We use $do(a_k=0)$ to denote the robot's inaction in dimension $i$, which means intervening on $a_k$ to set its value to 0. 
The main idea of our method is to compare the reward distributions when $a_k$ takes random value and when it is not executed while all other actions except $a_k$ are set to random values. This ensures that intervention only on $a_k$, while the random sampling of other actions can encompass more complex situations that may arise during task execution. Let $P_1 = P(r^i_j|\textbf{s},do(a_k=0))$ represent the probability distribution of $r^i_j$ after intervening on robot action $a_k$, $P_2 = P(r^i_j|\textbf{s},random(a_k))$ denote the probability distribution of $r^i_j$ when robot action $a_k$ takes a random value. If the difference between $P_1$ and $P_2$ is greater than a certain threshold, it is considered that $a_k$ is the causal variable of $r^i_j$, i.e., $a_k \rightarrow r^i_j$.

Below, we introduce the causal identification process within a single subtask $ST_i$.

\subsubsection{Collecting Data}\label{Collecting Data}
 For each $\textbf{r}^i$, placing the agent in the environment of $ST_i$ for interaction to obtain random data($D^i_{random}$) and intervention data($D^i_{do} = \{D_{do}^{i,a_1},\ldots,D_{do}^{i,a_K}\}$).
 $D^i_{random}$ refers to the data when action values in all dimensions take random values, while $D^i_{do}$ includes data when actions in each dimension are not executed individually. The data $D_{do}^{i,a_k},k \in (1,\ldots,K)$ for intervening on a single action 
$a_k$ must have the same number of data as $D^i_{random}$. 
At time $t$, both types of data include the environmental state $\textbf{s}_t$, the action taken $\textbf{a}_t$ and the reward $\textbf{r}^i_t$. If during the data collection process there is a situation where $u(\textbf{s}_t) \ne i$, it indicates that the agent has entered a different stage from i due to the execution of $\textbf{a}_t$, and therefore the environment needs to be reset to $ST_i$.
The random data $D^i_{random}$ and intervention data $D^i_{do}$ will then be combined for training the reward prediction model. In addition, it is necessary to collect a portion of data where $a_k$ takes random values ($D_{inference}^{i}$) for calculating the difference between the intervention and random probability distributions.

\subsubsection{Training Reward Prediction Models}
A separate reward prediction model is constructed for each pair $(a_k,r_j^i)$ in $ST_i$. We use a neural network to fit the reward situations under both random and intervention scenarios. For each action $a_k$, $D_{do}^{i,a_k}$ and $D_{random}^{i}$ are combined to train the neural network to get a reward prediction model. The reward prediction model takes the state of the environment $\textbf{s}$ and the action 
$\textbf{a}$ performed by the robot as input and output the mean and the logarithm of the standard deviation of the expected reward $\textbf{r}^i$. The mean and the standard deviation are then used to construct a normal distribution. The loss function is the negative log probability of $r_{j}^i$ under this normal distribution.

\subsubsection{Causal Discovery}
\begin{figure*}[htbp]
  \centering
    \includegraphics[scale=0.5]{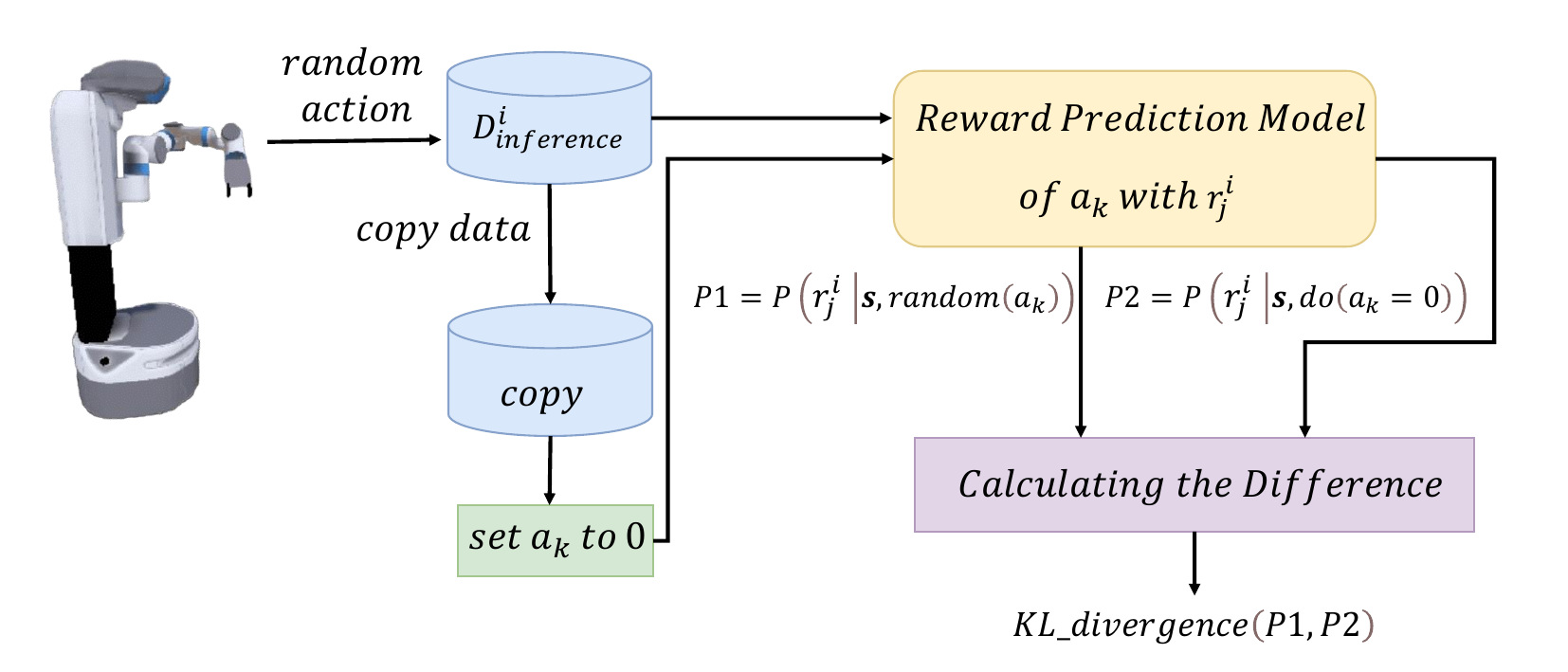}
   \caption{Procedure for Calculating the Difference for Each Pair($a_k$,$r_j^i$)}
  \label{inference}
\end{figure*}
After training the reward prediction model for each pair $(a_k,r_j^i)$ in $ST_i$, the distribution differences are calculated respectively.
For each pair $(a_k,r_j^i)$, we first create a copy of $D_{inference}^{i}$ and set all $a_k$ in the copy to 0. Setting the corresponding action in the copy to 0 simulates the scenario where action $a_k$ is not executed, while all other environmental states remain the same. Therefore, the \textbf{s} and $\textbf{a}$ in $D_{inference}^{i}$ and 
copy correspond to the data when the robot performs action $a_k$ or not in the same environment.
Then, we input the $\textbf{s}$ and $\textbf{a}$ from the copy and the $D_{inference}^{i}$ into the reward prediction model to get $P_1 = P(r_j^i|\textbf{s},do(a_k=0))$ and $P_2 = P(r_j^i|\textbf{s},random(a_k))$ separately. $P1$ and $P2$ are the reward distributions for taking or not taking $a_k$ under the same $\textbf{s}$. After obtaining $P1$ and $P2$, we need a criterion to measure the difference between these two distributions. KL divergence is commonly used to measure the distance between two probability distributions \cite{cover1999elements}. Therefore, we choose the KL divergence to calculate the difference between the distributions $P1$ and $P2$. 
The entire process of obtaining the KL divergence between the random distribution $P2$ and the intervention distribution $P1$ of $(a_k,r_j^i)$ is shown in the Fig.~\ref{inference}.

For $r_j^i$, a list of KL divergences that include all dimensions of actions will be obtained, represented as 
\begin{equation}
KLDlist_{r_j^i}=[kld_{r_j^i}^{a_1},kld_{r_j^i}^{a_2},...,kld_{r_j^i}^{a_K}]
 \end{equation}
The process of selecting causal actions for $r_j^i$ is illustrated in Algorithm \ref{alg:Selection of Causal Actions}. 
Assuming that there is always at least one action that affects $r_j^i$, we select the causal action based on the degree of difference in $KLDlist_{r_j^i}$.
The Coefficient of Variation is a normalized measure of the dispersion of a probability distribution, defined as the ratio of the standard deviation to the mean. 
Therefore, we use the coefficient of variation to measure the degree of difference in $KLDlist_{r_j^i}$. If the coefficient of variation of $KLDlist_{r_j^i}$ is greater than a threshold $\epsilon_{cv}$, this indicates that different actions have a significant difference in their impact on $r_j^i$. In this case, to avoid the influence of a particular action being too significant and overshadowing the effects of other causal actions, the $KLDlist_{r_j^i}$ needs to be normalized first. Therefore, normalize $KLDlist_{r_j^i}$ using min-max scaling to range between 0 and 1, and then select actions that have values greater than the threshold $\epsilon_{normalize}$ as the reasonable actions of the $r_j^i$.
If the coefficient of variation of $KLDlist_{r_j^i}$ is smaller than $\epsilon_{cv}$, it suggests that all actions have a similar impact on $r_j^i$ and none stand out significantly. In this cases, instead of performing min-max scaling on $KLDlist_{r_j^i}$, we directly select actions from it. Actions with KL divergence exceeding the threshold $\epsilon_{direct}$ are considered to be causal actions. 
After obtaining the set of causal actions for each $r_j^i$, it can be converted into the form of a causal matrix.

After multiple sets of tests, we found that setting $\epsilon_{cv}$ to 1.00, $\epsilon_{normalize}$ to 0.10, and $\epsilon_{direct}$ to 0.01 is more appropriate.
The entire process of causal discovery is shown in Algorithm \ref{alg:Causal Discovery}.

\begin{algorithm}
\caption{Causal Discovery}
\label{alg:Causal Discovery}
\begin{algorithmic}[1]
    \STATE \textbf{Input:}
    \STATE{\quad An agent capable of performing action \textbf{a}}
    \STATE{\quad $trainingCount$, number of training samples }
    \STATE{\quad $inferenceCount$, number of inference samples }
    \STATE{\quad Causal matrix list $causalMatrixs \leftarrow \{\}$}
    \STATE \textbf{Output:} Causal matrices for each stage
    \STATE
    \FOR{i from 1 to N}
        \STATE{$D^i_{random} \leftarrow \{\},D^i_{inference} \leftarrow \{\},D^i_{do} \leftarrow \{\}$}
        \WHILE{$collectedCount < trainingCount$}
            \STATE{collect random data add to $D^i_{random}$}
        \ENDWHILE
        \WHILE{$collectedCount < inferenceCount$}
            \STATE{collect predictive data add to $D^i_{inference}$}
        \ENDWHILE
        \FOR{$a \in \textbf{a}$}
            \WHILE{$collectedCount < trainingCount$}
                \STATE{collect intervention data on $a$ and add it to $D^i_{do}$}
            \ENDWHILE
        \ENDFOR
        \STATE{training reward prediction model of $\textbf{r}^i$ using $D^i_{random}$ and $D^i_{do}$}
        \STATE{obtain the causal matrix $m_i$ through the reward prediction model of $\textbf{r}^i$ and $D^i_{inference}$}
        \STATE{add $m_i$ to $causalMatrixs$}
    \ENDFOR
    \RETURN $causalMatrixs$
\end{algorithmic}  
\end{algorithm}

\begin{algorithm}
\caption{Selection of Causal Actions}
\label{alg:Selection of Causal Actions}
\begin{algorithmic}[1]
    \STATE \textbf{Input:}$KLDlist$, a list of KL divergences for a reward item 
    \STATE \textbf{Output:} $SelectedActions$, the causal actions for the reward item 
    \STATE
    \STATE{$\mu \leftarrow \text{mean}(KLDlist)$}
    \STATE{$\sigma \leftarrow \text{Standard\_Deviation}(KLDlist)$}
    \STATE{$cv \leftarrow \sigma/\mu$}
    \STATE{$SelectedActions \leftarrow \{\}$}
    \STATE{$threshold \leftarrow \epsilon_{direct}$}
    \IF{$cv > \epsilon_{cv}$}
        \STATE{$KLDlist \leftarrow \text{Min-Max\_Normalization}(KLDlist)$}
        \STATE{$threshold \leftarrow \epsilon_{normalize}$}
    \ENDIF
    \FOR{each $KLD \in KLDlist$}
        \IF{$KLD > threshold$}
            \STATE{$\text{add action corresponding to } KLD $}
            \STATE{$\text{to } SelectedActions$}
        \ENDIF
    \ENDFOR
    \RETURN $SelectedActions$
\end{algorithmic}  
\end{algorithm}

\subsection{Training}\label{Training}

\begin{algorithm}
\caption{Training}
\label{alg:Training}
\begin{algorithmic}[1]
    \STATE \textbf{Input:}
    \STATE{\quad $\{A_n|n \in (1,\ldots, n)\}$, set of N agents}
    \STATE{\quad $causalMatrixs$, causal matrix list }
    \STATE{\quad $totalSteps$, total training steps }
    \STATE{\quad $n\_step$, the number of steps sampled in an epoch }
    \STATE{\quad $count$, number of training samples }
    \STATE
    \FOR{i from 1 to N}
        \STATE{get Causal action $\textbf{ca}_i$ from $m_i$}
        \STATE{build $A_i$'s policy $\pi_i$ using $\textbf{ca}_i$}
    \ENDFOR
    \IF{The current algorithm is on-policy.}
        \WHILE{$currentSteps < totalSteps$}
            \FOR{step from 1 to $n\_step$}
                \STATE{Select agent $u(\textbf{s}_t)$ to interact with the environment, collecting state, action, and reward data.}
            \ENDFOR
            \FOR{i from 1 to N}
                \STATE{training $\pi_i$ with causal policy gradient method by using $m_i$}
            \ENDFOR
        \ENDWHILE
    \ELSE
        \WHILE{$currentSteps < totalSteps$}
            \FOR{step from 1 to $n\_step$}
                \STATE{Select agent $u(\textbf{s}_t)$ to interact with the environment, collecting state, action, and reward data.}
        
                \IF{The cumulative number of new samples obtained for agent $u(\textbf{s}_t) >= count$}
                    \STATE{training $\pi_i$ with causal policy gradient method by using $m_i$}
                \ENDIF
            \ENDFOR
        \ENDWHILE
    \ENDIF
\end{algorithmic}  
\end{algorithm}

The causal matrix uses action items as rows and reward items as columns. By following the causal discovery steps, we obtain causal matrixs $\{m_i|i \in {1,\ldots, n}\}$ corresponding to each stage's subtask. For causal matrix $m_i$, select the set of action terms whose row sums are not zero to construct the action space for agent $A_i$. In addition to using the causal actions to construct the action space for the policy to reduce redundant exploration in RL, the causal matrix $m_i$ is also integrated into the training process of RL using the causal policy gradient method. During the gradient ascent process, multiplying by the causal matrix $m_i$ ensures that only the reward values of the action outcomes are passed to the corresponding actions. This can effectively reduce the gradient variance during the training of each subtask. The training process is shown in algorithm \ref{alg:Training}.

\section{EXPERIMENTAL SETUP} \label{EXPERIMENTAL SETUP}

We conducted comparative experiments on a mobile manipulation task and a pure manipulation task.
The first type involves scenarios that require both robot mobility and manipulation by the robotic arm. The second type focuses solely on manipulation. 
Both tasks were conducted using the Fetch robot for experiments. 
The robot we use is equipped with a high-level controller, allowing the robot to directly perform macro actions such as moving forward and turning, and moving the end effector along the axes of its base coordinate system, without having to consider the control of each joint. Simulation experiments were conducted in iGibson \cite{li2022igibson}. 

\subsection{Mobile Manipulation Task}
\begin{figure}[htbp]
  \centering
  \begin{subfigure}[b]{0.4\linewidth}
    \includegraphics[height=0.15\textheight,width=\linewidth]{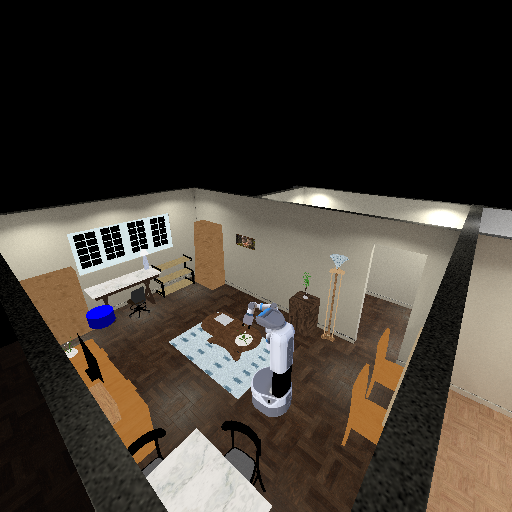}
    \caption{Stage 1}
    \label{mobile_manipulaiton_stage1}
  \end{subfigure}
  \begin{subfigure}[b]{0.4\linewidth}
    \includegraphics[height=0.15\textheight,width=\linewidth]{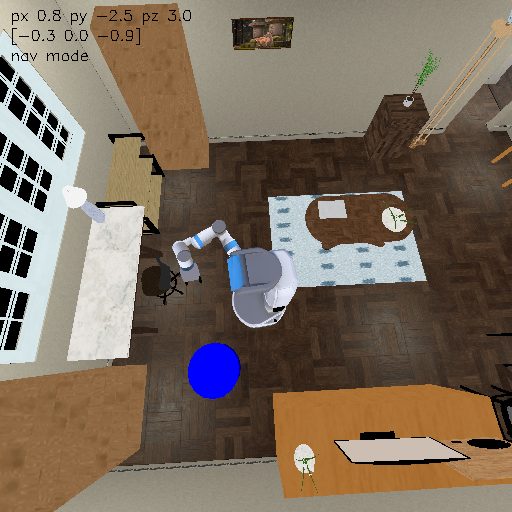}
    \caption{Stage 2}
    \label{mobile_manipulaiton_stage2}
  \end{subfigure}
    \begin{subfigure}[b]{0.4\linewidth}
    \includegraphics[height=0.15\textheight,width=\linewidth]{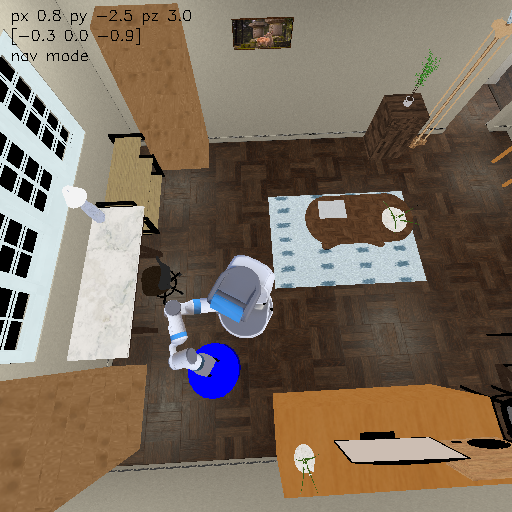}
    \caption{Stage 3}
    \label{mobile_manipulaiton_stage3}
  \end{subfigure}
  \caption{Mobile Manipulation Task}
  \label{task1}
\end{figure}

The mobile manipulation task scenario is set in an indoor living room, where each time randomly generates a target point. The robot needs to navigate autonomously to the vicinity of the target point and place its end effector inside the target point.
The mobile manipulation task consists of three stages:
\begin{enumerate}
    \item Navigate to a certain range around the target point. This stage includes a reward item $r_\rho$ that measures the horizontal distance from the robot's base to the target point.
    \item The robot has reached near the target point and needs to turn to face the target point directly. This stage includes a reward item $r_\theta$ that measures the angle between the robot and the target point.
    \item The robot, already near the target point and facing it,
     moves the end effector of the manipulator to within the
     target point. This stage includes 3 reward items $r_{eefx},r_{eefy},r_{eefz}$, each measuring the distance between the end effector and the target point along the axes of the world coordinate system.
\end{enumerate}
The stages of the task are shown in the Fig.\ref{task1}.
In addition to the aforementioned reward items, the mobile manipulation task also has three reward items: base collision, robotic arm collision, and self collision. As collision-related reward items do not signify task success, these three are not utilized in the causal discovery phase, but are only used for training. During training, the corresponding positions in the causal matrix for the reward items of these three types of collisions are all set to 1.

In this task, the Fetch robot is capable of moving forward $forward$, turning $turn$, and moving the end effector along its own base coordinate system $armx,amry,armz$, totaling 5 executable actions. The observation space for this task is detailed in the appendix \ref{mobile manipulation observation space}.

We conducted comparative experiments based on the on-policy RL method PPO \cite{schulman2017proximal} for this task. Our method is named cmPPO (causal multiple PPO). 
In cmPPO, the action space of each agent is composed of causal actions obtained through causal discovery, and causal policy gradient learning is implemented using the corresponding causal matrix. It is compared to the regular PPO algorithm (mPPO). In mPPO, each subtask utilizes all available actions as its action space, and each sub-policy is trained using the standard PPO algorithm.

\subsection{Pure Manipulation Task}
\begin{figure}[htbp]
  \centering
  \begin{subfigure}[b]{0.4\linewidth}
    \includegraphics[width=\linewidth]{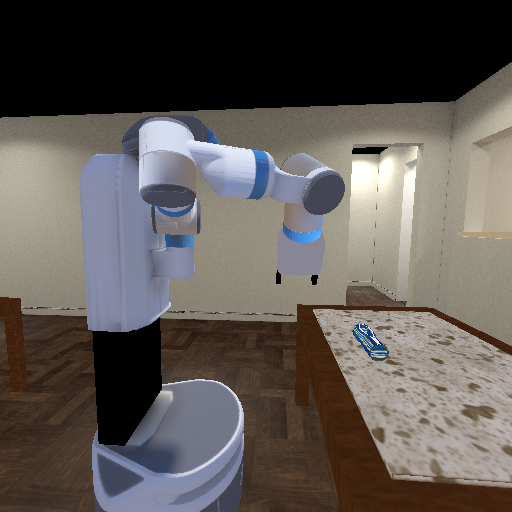}
    \caption{Stage 1}
    \label{grasp_task_stage1}
  \end{subfigure}
  \quad
  \quad
  \begin{subfigure}[b]{0.4\linewidth}
    \includegraphics[width=\linewidth]{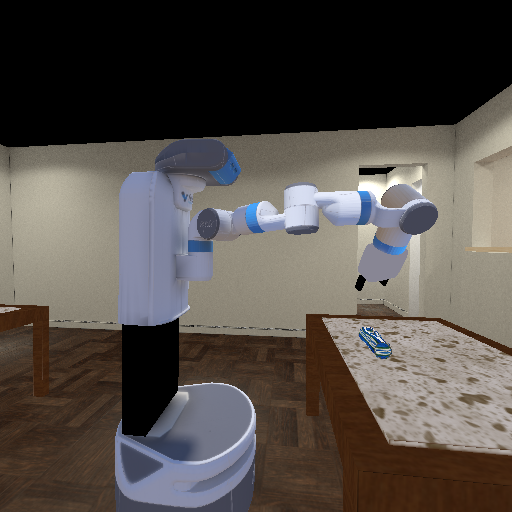}
    \caption{Stage 2}
    \label{grasp_task_stage2}
  \end{subfigure}
  
  \begin{subfigure}[b]{0.4\linewidth}
    \includegraphics[width=\linewidth]{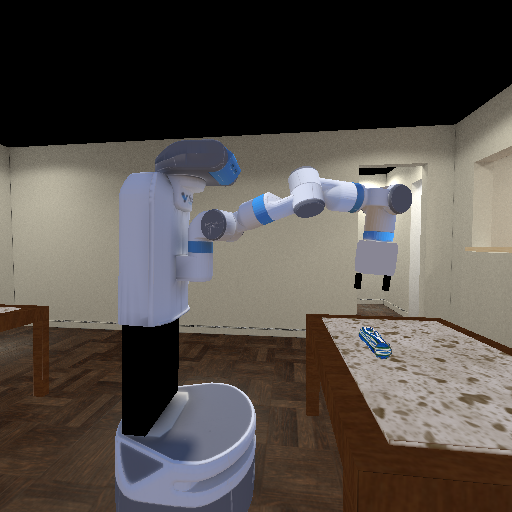}
    \caption{Stage 3}
    \label{grasp_task_stage3}
  \end{subfigure}
  \quad
  \quad
  \begin{subfigure}[b]{0.4\linewidth}
    \includegraphics[width=\linewidth]{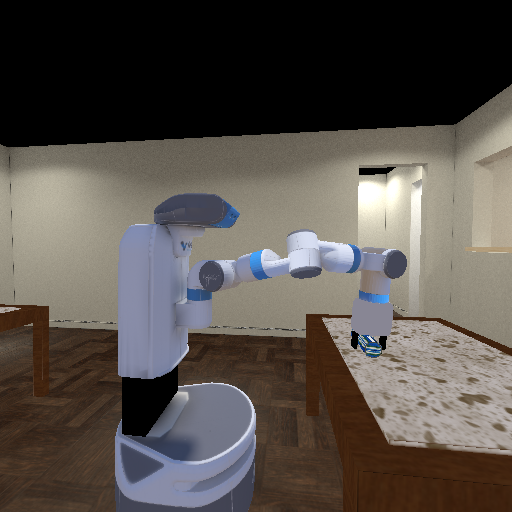}
    \caption{Stage 4}
    \label{grasp_task_stage4}
  \end{subfigure}
  \caption{Pure Manipulation Task}
  \label{task2}
\end{figure}

This task is a four-stage mission that requires the Fetch robot to use its end effector to grasp a pencil box placed on a table:
\begin{enumerate}
    \item Move the end effector to a certain height directly above the pencil box. This stage involves three reward items $r_{eefx},r_{eefy},r_{eefz1}$, which measure the distance between the end effector and the target point above the pencil box along the three axes of the world coordinate system.
    \item Maintain the position of the end effector reached in stage 1, and rotate the end effector to face downward. This stage includes a reward item $r_{ori}$ measuring the distance between the current pose of the end effector and its pose when facing downward.
    \item The end effector maintains the posture reached in stage 2, and stays at the horizontal position achieved in stage 1, only lowering the height of the end effector until it can grasp the pencil box. This stage involves a reward component $r_{eefz2}$ that measures the height difference between the end effector and the pencil box.
    \item Close the end effector to grasp the pencil box. This stage involves a reward item $r_{gripper}$
\end{enumerate}
The stages of the task are shown in the Fig.\ref{task2}.
Stages 1 to 3 also include the reward $r_{gripper}$ in addition to the aforementioned rewards. This is because the end effector should not close before reaching stage 4. Similarly, the rewards for each stage also take into account collisions of the robotic arm and self collisions. These two reward items do not participate in the causal discovery step, but are directly applied to the training step.

In this task the Fetch robot has executable actions such as moving the end effector along its own base coordinate system $armx,army,armz$, rotating around the end effector coordinate system $armrx,armry,armrz$, and grasping $grip$. The observation space for this task is detailed in the appendix \ref{pure manipulation observation space}.

For this task, we conducted comparative experiments based on the off-policy RL algorithm Soft Actor-Critic (SAC) \cite{haarnoja2018soft}. Our proposed method is called cmSAC (causal multiple SAC). Similarly to mPPO, this experiment also sets up mSAC(multiple SAC) for comparison. Furthermore, to compare the differences in RL training between manually specified causal relationships and those obtained using automatic discovery methods, the cmmSAC (causal manual multiple SAC) method was introduced for this task. This method uses a manually specified causal graph to construct the action space and perform causal policy gradient learning. Finally, the cSAC (cooperative SAC) method \cite{erskine2022developing}, which involves cooperative learning among agents, was also employed. In this method, each sub-agent similarly uses all actions to construct the action space.

\subsection{Reward Setting}
The reward setting logic for the two tasks is the same.
Let $x^{t}$ denote the value of one of the environmental variables $X$ at time $t$, and $x^{target}$ represent the value this environmental variable should take when the subtask is completed.
$D^{t} = \left\| x^{t} - x^{target} \right\|$ represent the distance to the target point. For example, in mobile manipulation task, $D^{t}$ can represent the horizontal distance between the robot base and the target point at time $t$. The reward at time $t$ is defined as $r^t = \lambda (D^{t-1} - D^{t})$, $\lambda$ is a constant. This reward encourages agents to adjust their responsible state variables to appropriate values. For the reward item related to collisions, the value is usually 0, and a fixed negative reward is given once a collision occurs. For the reward 
$r_{gripper}$, a fixed negative reward is given if the end effector closes before stage 4 of the pure manipulation task. If it successfully grasps the pencil box by closing in stage 4, a fixed positive reward is given. In both tasks, a fixed reward value is directly given upon successful completion of a subtask.

The neural network structures and hyperparameter settings used in the experiments are detailed in the appendix \ref{neural network architecture}, \ref{Hyperparameter setting}. 

\section{RESULT} \label{RESULT}
\begin{table*}[ht]  
  \caption{mobile manipulation task KL divergence}  
  \renewcommand{\arraystretch}{2}
  \setlength{\tabcolsep}{3pt}
  \begin{center}
  \begin{tabular}{|c|c|c|c|c|c|c|c|c|c|c|c|c|}  
    \hline  
    \textbf{stage} & \textbf{reward} &\textbf{forward} & \textbf{turn} & \textbf{armx} & \textbf{army} & \textbf{armz} & \textbf{cv} &
    \textbf{forward\_n} & \textbf{turn\_n} & \textbf{armx\_n} & \textbf{army\_n} & \textbf{armz\_n}
    \\
    \hline
    \multirow{1}{*}{1}
        &   $\textbf{r}_{\rho}$
        &   0.0532 
        &   0.0122
        &   0.0002
        &   0.0003
        &   0.0001
        &   1.74
        &   1.0000
        &   0.2279
        &   0.0012
        &   0.0033
        &   0.0000
        \\
    \cline{1-13}
    \hline
    \multirow{1}{*}{2}
        &   $\textbf{r}_{\theta}$
        &   0.1131 
        &   0.1056
        &   0.0050
        &   0.0007
        &   0.0001
        &   1.17
        &   1.0000
        &   0.9343
        &   0.0438
        &   0.0054
        &   0.0000
        \\
    \cline{1-13}
    \multirow{3}{*}{3}
        &   $\textbf{r}_{eefx}$
        &   0.0076
        &   0.0163
        &   0.3679
        &   1.3100
        &   0.0236
        &   1.62
        &   0.0000
        &   0.0067
        &   0.2767
        &   1.0000
        &   0.0123
        \\
        \cline{2-13}
        &   $\textbf{r}_{eefy}$
        &   0.0084
        &   0.0042
        &   1.6487
        &   0.2820
        &   0.0167
        &   1.82
        &   0.0025
        &   0.0000
        &   1.0000
        &   0.1689
        &   0.0076
        \\
        \cline{2-13}
        &   $\textbf{r}_{eefz}$
        &   0.0017
        &   0.0007
        &   0.2389
        &   0.0128
        &   8.7254
        &   2.16
        &   0.0001
        &   0.0000
        &   0.0273
        &   0.0014
        &   1.0000
        \\
    \cline{1-13}
  \end{tabular}
  \label{tab:task1_amplitude1}  
  \end{center}
\end{table*}

\begin{table*}[ht]  
  \caption{pure manipulation task KL divergence}  
  \renewcommand{\arraystretch}{2}
  \setlength{\tabcolsep}{2.5pt}
  \begin{center}
  \begin{tabular}{|c|c|c|c|c|c|c|c|c|c|c|c|c|c|c|c|c|}  
    \hline  
    \textbf{stage} & \textbf{reward} & \textbf{armx} & \textbf{army} & \textbf{armz} 
    & \textbf{armrx} & \textbf{armry} & \textbf{armrz} & \textbf{grip}
    & \textbf{cv} 
    & \textbf{armx\_n} & \textbf{army\_n} & \textbf{armz\_n} 
    & \textbf{armrx\_n} & \textbf{armry\_n} & \textbf{armrz\_n} 
     & \textbf{grip\_n}
    \\
    \hline
    \multirow{3}{*}{1}
        &   $\textbf{r}_{eefx}$
        &   0.0001
        &   0.3381
        &   0.0003
        &   0.0001
        &   0.0000
        &   0.0009
        &   0.0001
        &   2.44
        &   0.0002
        &   1.0000
        &   0.0008
        &   0.0003
        &   0.0000
        &   0.0026
        &   0.0002
        \\
        \cline{2-17}
        &   $\textbf{r}_{eefy}$
        &   0.1080
        &   0.0000
        &   0.0003
        &   0.0000
        &   0.0012
        &   0.0011
        &   0.0000
        &   2.38
        &   1.0000
        &   0.0003
        &   0.0030
        &   0.0000
        &   0.0107
        &   0.0103
        &   0.0000
        \\
        \cline{2-17}
        &   $\textbf{r}_{eefz1}$
        &   0.0071
        &   0.0000
        &   0.0456
        &   0.0009
        &   0.0009
        &   0.0005
        &   0.0001
        &   1.98
        &   0.1561
        &   0.0000
        &   1.0000
        &   0.0190
        &   0.0188
        &   0.0100
        &   0.0010
        \\
    \cline{1-17}
    \multirow{1}{*}{2}
        &   $\textbf{r}_{ori}$
        &   0.0045
        &   0.0007
        &   0.0011
        &   0.2246
        &   0.0599
        &   0.0130
        &   0.0006
        &   1.76
        &   0.0173
        &   0.0002
        &   0.0022
        &   1.0000
        &   0.2648
        &   0.0550
        &   0.0000
        \\
    \cline{1-17}
    \hline
    \multirow{1}{*}{3}
        &   $\textbf{r}_{eefz2}$
        &   0.0717
        &   0.0018
        &   3.4506
        &   0.0658
        &   0.0550
        &   0.0149
        &   0.0012
        &   2.29
        &   0.0204
        &   0.0002
        &   1.0000
        &   0.0187
        &   0.0156
        &   0.0040
        &   0.0000
        \\
    \cline{1-17}
    \hline
    \multirow{1}{*}{4}
        &   $\textbf{r}_{gripper}$
        &   0.0107
        &   0.0025
        &   0.0061
        &   0.0001
        &   0.0107
        &   0.0001
        &   4.6423
        &   2.43
        &   0.0023
        &   0.0005
        &   0.0013
        &   0.0000
        &   0.0023
        &   0.0000
        &   1.0000
        \\
    \cline{1-17}
  \end{tabular}
  \label{tab:task2}  
  \end{center}
\end{table*}

\begin{figure}[]
	\centerline{\includegraphics[scale=0.6]{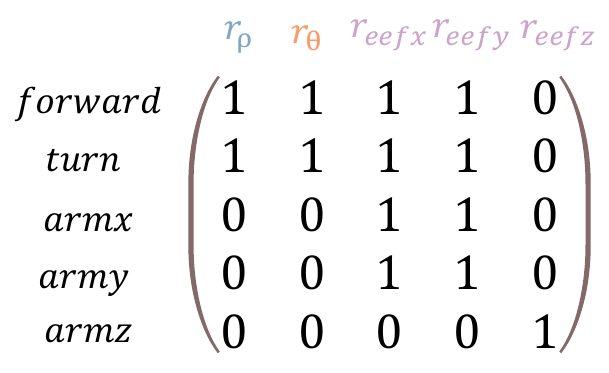}}
	\caption{Causal Matrix 1}
	\label{Causal Matrix 1}
\end{figure}
\begin{figure}[]
	\centerline{\includegraphics[scale=0.6]{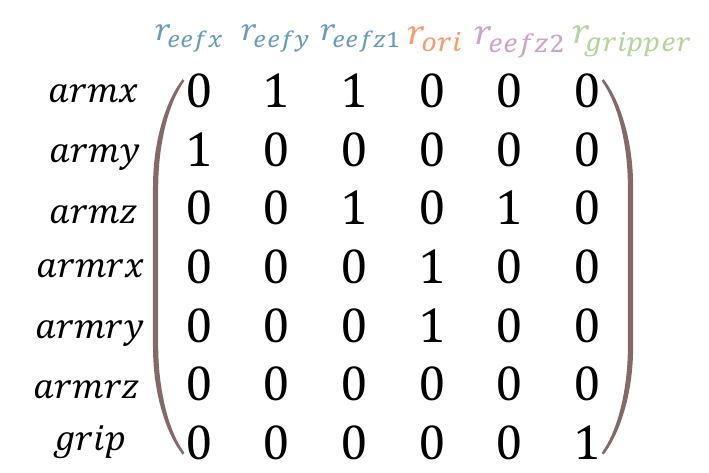}}
	\caption{Causal Matrix 2}
	\label{Causal Matrix 2}
\end{figure}
\begin{figure}[]
	\centerline{\includegraphics[scale=0.5]{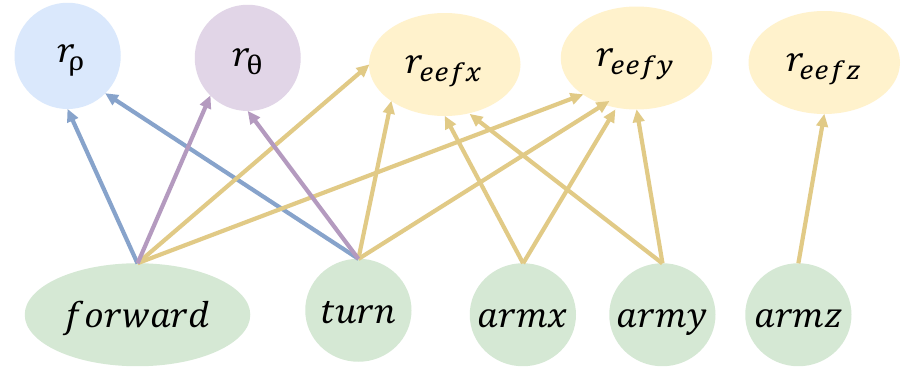}}
	\caption{Causal Graph 1}
	\label{Causal Graph 1}
\end{figure}
\begin{figure}[]
	\centerline{\includegraphics[width=9cm,height=3cm]{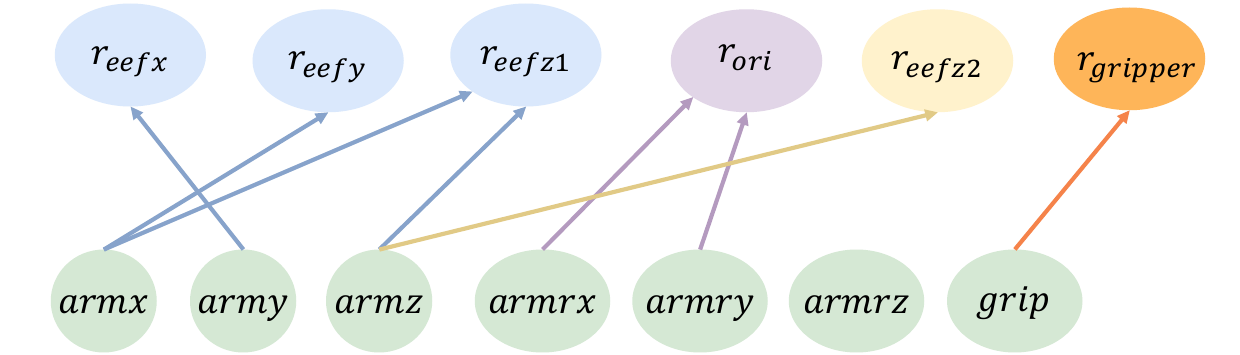}}
	\caption{Causal Graph 2}
	\label{Causal Graph 2}
\end{figure}

The causal matrices obtained from the two tasks are shown in Fig. \ref{Causal Matrix 1} and Fig. \ref{Causal Matrix 2}. For the sake of concise representation in this paper, we depict the causal matrices of multiple stages of a task as a single matrix, and use different colors to label the reward items to distinguish between task stages.
The KL divergence, coefficient of variation, and the results of the normalized KL divergence for the mobile manipulation task and the pure operation task are shown in Table 1 and Table 2, respectively. Here, $\textbf{cv}$ denotes the coefficient of variation, and variables with the suffix "$\textbf{n}$" correspond to the normalized values. To more intuitively observe the causal relationships between actions and reward items, the causal matrices are drawn into the causal graphs shown in Fig.\ref{Causal Graph 1} and Fig.\ref{Causal Graph 2}. Similarly, for ease of presentation, the causal graphs for different stages are drawn together, with reward items marked in different colors. Except for $r_{gripper}$, different colors represent rewards of different stages. $r_{gripper}$ is a reward item that exists in every stage of the pure manipulation task.

It is worth noting that in stage 3 of the mobile manipulation task, $forward$ and $turn$ are also causal actions. Because under the default settings of the iGibson controller, the magnitude of change for these two actions is relatively large compared to the actions of moving the robotic arm. Although performing a significant degree of movement in stage 3 can cause the stage to regress to stage 2 or 1, making small forward movements within a limited range will not. Moreover, due to the noticeable magnitude of movement of these two actions, their impact on the horizontal position of the end effector in stage 3 is not less than that of the actions of moving the robotic arm.

\begin{figure}[]
	\centerline{\includegraphics[width=9cm,height=3cm]{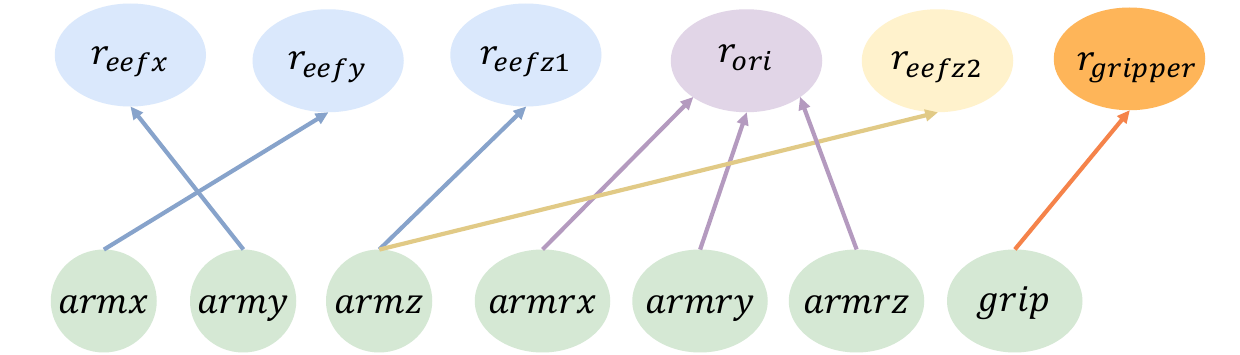}}
	\caption{Manual Causal Graph}
	\label{Manual Causal Graph}
\end{figure}

In the scenario setting of the grasping task, the robot base coordinate system is exactly orthogonal to the world coordinate system. The reward items are defined relative to the world coordinate system, while the end effector movement actions of the robotic arm are defined relative to the robot base coordinate system. Therefore, the results indicate that $armx$ affects $r_{eefy}$, and $army$ affects $r_{eefx}$.
It is worth noting that in the pure manipulation task, moving the end effector along the robot's own coordinate system x-axis will also affect the reward of the end effector on the z-axis. This is because the length of the robotic arm is limited. In order to continue moving forward along the x-axis, it will force the end effector to lower its height. Because the robot has a macro controller that can directly move the end effector along its own coordinate system, if humans manually provide the causal relationship, it may mistakenly think that $armx$ is not the causal action of $r_{eefz1}$. As for $r_{eefz2}$, once the end effector moves horizontally beyond the specified range leading to $u(\textbf{s}_t) \ne u(\textbf{s}_{t-1})$, it is no longer in the stage of $r_{eefz2}$. Therefore, $armx$ is not a causal variable of $r_{eefz2}$. It was also found that $armrz$ does not affect the posture reward $r_{ori}$, as the initial pose of the end effector only needs to be adjusted around the x and y axes to achieve the desired pose. If the causal graphs for these two cases are manually provided, then $armx$ would not be a cause of $r_{eefz1}$, while $armrz$ would be a cause of $r_{ori}$. The manually proposed causal relationships are shown in the Fig.\ref{Manual Causal Graph}. Experiments with cmmSAC were conducted based on this manually provided causal graph. 

According to the obtained causal relationships, the causal action sets used for constructing the action spaces at each stage are shown in the table \ref{tab:action space}. $grip$ is the causal action of $r_{gripper}$, and the reward $r_{gripper}$ exists in every stage of the pure manipulation task. Therefore, the action space of every stage of the pure manipulation task includes $grip$.

\begin{table}[ht]  
  \caption{Action Spaces}  
  \renewcommand{\arraystretch}{2}
  \setlength{\tabcolsep}{4pt}
  \begin{center}
  \begin{tabular}{|c|c|c|}  
    \hline  
    \textbf{task} & \textbf{stage} &\textbf{actions}
    \\
    \hline
    \multirow{1}{*}{mobile manipulation}
        & 1
        & $forward,turn$
        \\
        \cline{2-3}
        & 2
        & $forward,turn$
        \\
        \cline{2-3}
        & 3
        & $forward,turn,armx,army,armz$
        \\
    \cline{1-3}
    \multirow{3}{*}{pure manipulation}
        &   1
        &   $armx,army,armz,grip$
        \\
        \cline{2-3}
        &   2
        &   $armrx,armry,grip$
        \\
        \cline{2-3}
        &   3
        &   $armz,grip$
        \\
        \cline{2-3}
        &   4
        &   $grip$
        \\
        
    \cline{1-3}
  \end{tabular}
  \label{tab:action space}  
  \end{center}
\end{table}

\begin{figure}[]
	\centerline{\includegraphics[scale=0.5]{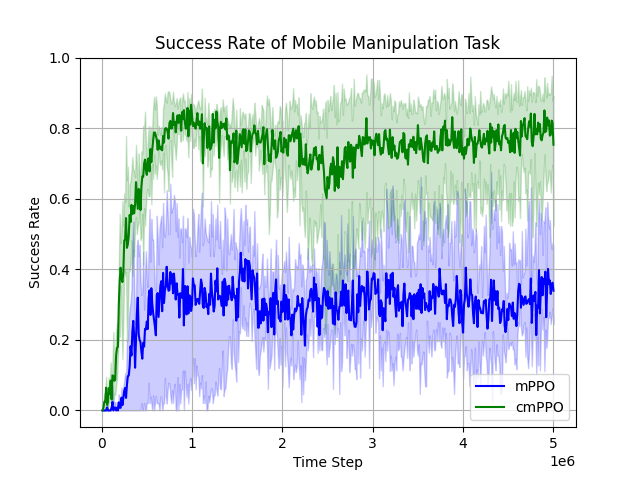}}
	\caption{Mobile Manipulation Task}
	\label{Success Rate:Mobile Manipulation Task}
\end{figure}

\begin{figure}[]
	\centerline{\includegraphics[scale=0.5]{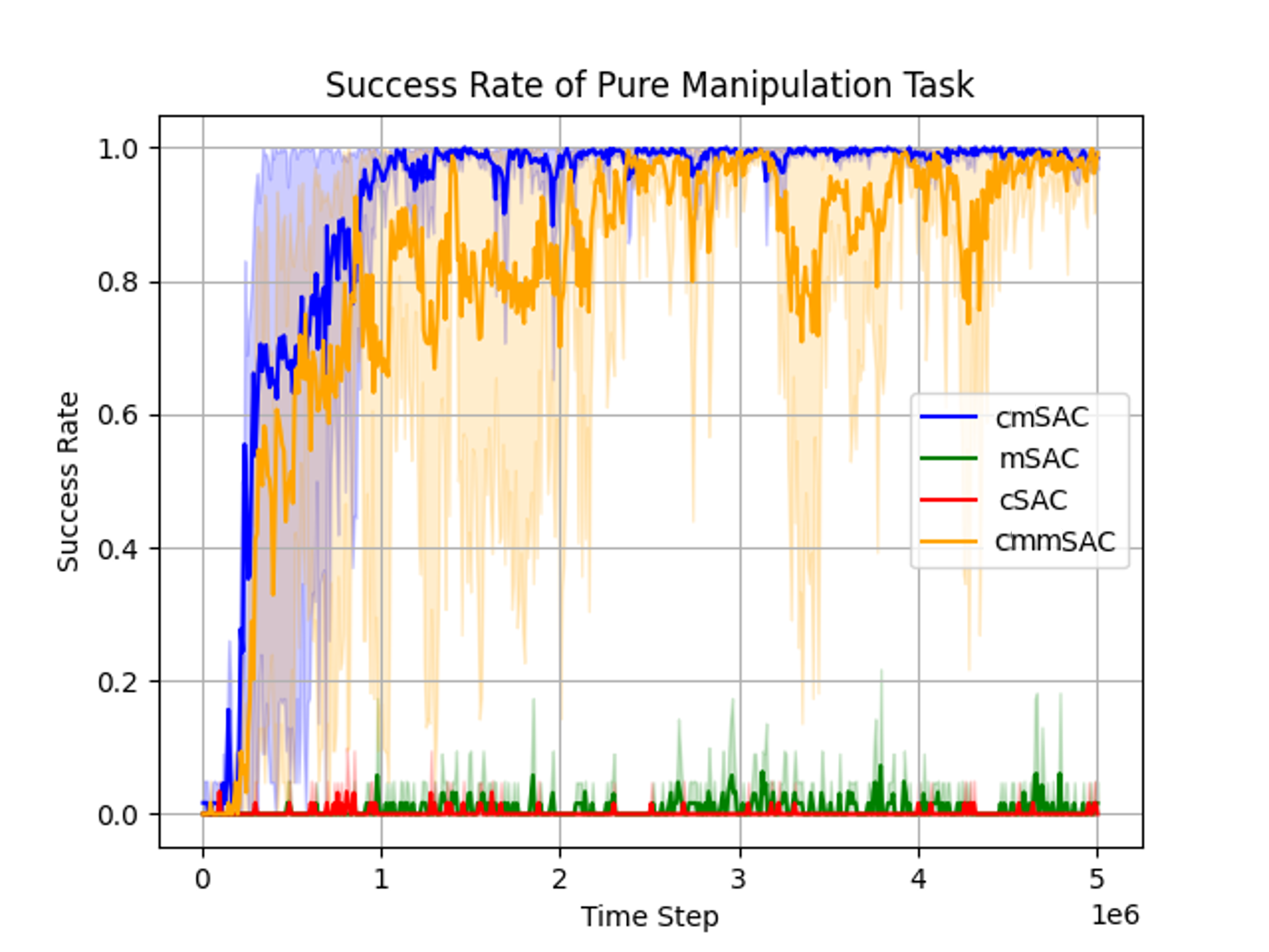}}
	\caption{Pure Manipulation Task}
	\label{Success Rate:Pure Manipulation Task}
\end{figure}

For the mobile manipulation task, comparative experiments are conducted using the causal relationships obtained above and based on the PPO algorithm for RL. The success rate curve of the mobile manipulation task is shown in the Fig.\ref{Success Rate:Mobile Manipulation Task}. It can be seen that our cmPPO method has a significantly higher success rate than the mPPO method under the same number of time steps. Here, it can be preliminarily demonstrated that using causal actions and causal policy gradient methods can effectively enhance the performance of RL. For the pure manipulation task, experiments are conducted based on the off-policy SAC method. The success rate curves of the four methods based on SAC are shown in the Fig.\ref{Success Rate:Pure Manipulation Task}. In the experimental results of this task, it can be seen that the two methods using causal actions and causal policy gradients, cmSAC and cmmSAC, can complete the task effectively. However, the cSAC and mSAC methods, which use all available actions to construct the action space, show difficulty in completing this four-stage task. The performance of the latter two methods often involves moving from stage 1 to stage 2 and then regressing back to stage 1, resulting in a reversal of progress. Due to frequent reversals of progress caused by actions in other dimensions during the learning process, the entire learning task struggles to advance to later stages. The cSAC method allows tasks between different stages to learn from each other through cooperation. However, due to a large number of reversals in progress, the policies of the subtasks in the later stages are difficult to learn, leading to poor performance of the cSAC method on this task. By using causal actions, it is possible to eliminate to the greatest extent the unnecessary actions for completing subtasks. This not only reduces redundant exploration but also helps to avoid some cases of progress reversal.

For cmSAC and cmmSAC, they both use causal actions and causal policy gradient methods. The former uses the causal relationships obtained by the automatic discovery method mentioned above, while the latter uses the causal relationships proposed by humans. 
From the success rate curve, it can be seen that at the same number of time steps, the cmSAC method has a more stable and higher success rate than cmmSAC. This also indicates that the causal relationships discovered through the robot's own interactions in the environment are more accurate than those specified by humans. Causal relationships obtained through interaction in specific environments can reveal points that humans might overlook. Utilizing this more precise causal relationship can lead to better performance in deep RL. 

\section{Conclusion} \label{section5}
In this work, we propose a method that utilizes causal relationships to handle robotic multi-stage tasks. Multi-stage tasks are broken down into multiple separate subtasks and addressed from the perspective of deep RL. By enabling the robot to interact with the environment, the causal relationships between the robot's macro actions and reward items are automatically discovered. The discovered causal actions are used to construct the action space of the agent corresponding to each subtask. Moreover, the causal relationships are integrated into the learning process of the subtasks using the causal policy gradient method. Experiments on a mobile manipulation task and a pure manipulation task demonstrate that the automatic discovery method, which leverages the robot's interaction with the environment, can obtain more accurate causal relationships than those specified by humans. Applying more accurate causal relationships to the causal policy gradient method can achieve better performance in RL. Furthermore, constructing the action space using causal actions can reduce redundant exploration in the RL process and also help to avoid some cases of progress reversal in multi-stage tasks.

\section{Limitations} \label{section6}
Although the method proposed in this paper significantly improves the performance of RL in multi-stage robotic tasks through causal relationships, there are still some limitations in terms of the generalization ability of causal relationships, the applicability to multi-robot collaborative tasks, and the dependence on task decomposition. Firstly, the discovery of causal relationships in this paper relies on the interaction data between the robot and the environment, which means that these causal relationships need to be relearned under different environmental conditions or task variants. Secondly, the method in this paper mainly focuses on the causal relationships between individual robot actions and rewards. In scenarios involving multiple robots collaborating to complete a task, the actions of individual robots may have complex causal dependencies with the actions of other robots, and the method proposed in this paper may not be able to effectively handle such situations. Moreover, the effectiveness of this method depends on whether the task can be reasonably decomposed into multiple subtasks. If the task decomposition is not appropriate, it may lead to the identification of incorrect causal actions, making the task difficult to complete.
Future work can explore how to enhance the generalizability of causal relationships, extend the method to handle causal relationships in multi-robot collaborative tasks, and investigate methods for reasonable task decomposition, thereby further improving the applicability of this method in diverse task scenarios.

\bibliographystyle{ieeetr}

\begin{thebibliography}{10}

\bibitem{duguleana2016neural}
M.~Duguleana and G.~Mogan, ``Neural networks based reinforcement learning for mobile robots obstacle avoidance,'' {\em Expert Systems with Applications}, vol.~62, pp.~104--115, 2016.

\bibitem{kulhanek2019vision}
J.~Kulh{\'a}nek, E.~Derner, T.~De~Bruin, and R.~Babu{\v{s}}ka, ``Vision-based navigation using deep reinforcement learning,'' in {\em 2019 european conference on mobile robots (ECMR)}, pp.~1--8, IEEE, 2019.

\bibitem{gu2017deep}
S.~Gu, E.~Holly, T.~Lillicrap, and S.~Levine, ``Deep reinforcement learning for robotic manipulation with asynchronous off-policy updates,'' in {\em 2017 IEEE international conference on robotics and automation (ICRA)}, pp.~3389--3396, IEEE, 2017.

\bibitem{andrychowicz2017hindsight}
M.~Andrychowicz, F.~Wolski, A.~Ray, J.~Schneider, R.~Fong, P.~Welinder, B.~McGrew, J.~Tobin, O.~Pieter~Abbeel, and W.~Zaremba, ``Hindsight experience replay,'' {\em Advances in neural information processing systems}, vol.~30, 2017.

\bibitem{wang2020learning}
C.~Wang, Q.~Zhang, Q.~Tian, S.~Li, X.~Wang, D.~Lane, Y.~Petillot, and S.~Wang, ``Learning mobile manipulation through deep reinforcement learning,'' {\em Sensors}, vol.~20, no.~3, p.~939, 2020.

\bibitem{hundt2020good}
A.~Hundt, B.~Killeen, N.~Greene, H.~Wu, H.~Kwon, C.~Paxton, and G.~D. Hager, ``“good robot!”: Efficient reinforcement learning for multi-step visual tasks with sim to real transfer,'' {\em IEEE Robotics and Automation Letters}, vol.~5, no.~4, pp.~6724--6731, 2020.

\bibitem{hu2023causal}
J.~Hu, P.~Stone, and R.~Mart{\'\i}n-Mart{\'\i}n, ``Causal policy gradient for whole-body mobile manipulation,'' {\em arXiv preprint arXiv:2305.04866}, 2023.

\bibitem{stengel2022guiding}
E.~Stengel-Eskin, A.~Hundt, Z.~He, A.~Murali, N.~Gopalan, M.~Gombolay, and G.~Hager, ``Guiding multi-step rearrangement tasks with natural language instructions,'' in {\em Conference on Robot Learning}, pp.~1486--1501, PMLR, 2022.

\bibitem{bao2022learn}
J.~Bao, G.~Zhang, Y.~Peng, Z.~Shao, and A.~Song, ``Learn multi-step object sorting tasks through deep reinforcement learning,'' {\em Robotica}, vol.~40, no.~11, pp.~3878--3894, 2022.

\bibitem{oh2017zero}
J.~Oh, S.~Singh, H.~Lee, and P.~Kohli, ``Zero-shot task generalization with multi-task deep reinforcement learning,'' in {\em International Conference on Machine Learning}, pp.~2661--2670, PMLR, 2017.

\bibitem{nachum2018data}
O.~Nachum, S.~S. Gu, H.~Lee, and S.~Levine, ``Data-efficient hierarchical reinforcement learning,'' {\em Advances in neural information processing systems}, vol.~31, 2018.

\bibitem{li2020hrl4in}
C.~Li, F.~Xia, R.~Martin-Martin, and S.~Savarese, ``Hrl4in: Hierarchical reinforcement learning for interactive navigation with mobile manipulators,'' in {\em Conference on Robot Learning}, pp.~603--616, PMLR, 2020.

\bibitem{luo2022relay}
Y.~Luo, Y.~Wang, K.~Dong, Q.~Zhang, E.~Cheng, Z.~Sun, and B.~Song, ``Relay hindsight experience replay: Continual reinforcement learning for robot manipulation tasks with sparse rewards,'' {\em arXiv preprint arXiv:2208.00843}, 2022.

\bibitem{erskine2022developing}
J.~Erskine and C.~Lehnert, ``Developing cooperative policies for multi-stage reinforcement learning tasks,'' {\em IEEE Robotics and Automation Letters}, vol.~7, no.~3, pp.~6590--6597, 2022.

\bibitem{wang2023multi}
D.~Wang, F.~Chang, and C.~Liu, ``Multi-stage reinforcement learning for non-prehensile manipulation,'' {\em arXiv preprint arXiv:2307.12074}, 2023.

\bibitem{andreas2017modular}
J.~Andreas, D.~Klein, and S.~Levine, ``Modular multitask reinforcement learning with policy sketches,'' in {\em International conference on machine learning}, pp.~166--175, PMLR, 2017.

\bibitem{gasse2021causal}
M.~Gasse, D.~Grasset, G.~Gaudron, and P.-Y. Oudeyer, ``Causal reinforcement learning using observational and interventional data,'' {\em arXiv preprint arXiv:2106.14421}, 2021.

\bibitem{pearl2012calculus}
J.~Pearl, ``The do-calculus revisited,'' {\em arXiv preprint arXiv:1210.4852}, 2012.

\bibitem{hu2022causality}
X.~Hu, R.~Zhang, K.~Tang, J.~Guo, Q.~Yi, R.~Chen, Z.~Du, L.~Li, Q.~Guo, Y.~Chen, {\em et~al.}, ``Causality-driven hierarchical structure discovery for reinforcement learning,'' {\em Advances in Neural Information Processing Systems}, vol.~35, pp.~20064--20076, 2022.

\bibitem{pearl2016causal}
J.~Pearl, M.~Glymour, and N.~P. Jewell, {\em Causal inference in statistics: A primer}.
\newblock John Wiley \& Sons, 2016.

\bibitem{pearl2009causality}
J.~Pearl, {\em Causality}.
\newblock Cambridge university press, 2009.

\bibitem{cover1999elements}
T.~M. Cover, {\em Elements of information theory}.
\newblock John Wiley \& Sons, 1999.

\bibitem{li2022igibson}
C.~Li, F.~Xia, R.~Mart\'in-Mart\'in, M.~Lingelbach, S.~Srivastava, B.~Shen, K.~E. Vainio, C.~Gokmen, G.~Dharan, T.~Jain, A.~Kurenkov, K.~Liu, H.~Gweon, J.~Wu, L.~Fei-Fei, and S.~Savarese, ``igibson 2.0: Object-centric simulation for robot learning of everyday household tasks,'' in {\em Proceedings of the 5th Conference on Robot Learning} (A.~Faust, D.~Hsu, and G.~Neumann, eds.), vol.~164 of {\em Proceedings of Machine Learning Research}, pp.~455--465, PMLR, 08--11 Nov 2022.

\bibitem{schulman2017proximal}
J.~Schulman, F.~Wolski, P.~Dhariwal, A.~Radford, and O.~Klimov, ``Proximal policy optimization algorithms,'' {\em arXiv preprint arXiv:1707.06347}, 2017.

\bibitem{haarnoja2018soft}
T.~Haarnoja, A.~Zhou, P.~Abbeel, and S.~Levine, ``Soft actor-critic: Off-policy maximum entropy deep reinforcement learning with a stochastic actor,'' in {\em International conference on machine learning}, pp.~1861--1870, PMLR, 2018.

\end{thebibliography}

\appendix
\section{appendix}
\subsection{Mobile Manipulation Observation Space}
\label{mobile manipulation observation space}
The observable state variables of the robot in this task include:
\begin{enumerate}
    \item $\rho$ : the polar radius of the target point relative to the robot's base coordinate system in the horizontal plane,
    \item $\theta$ : the polar angle of the target point relative to the robot's base coordinate system in the horizontal plane,
    \item $height$ : the height of the target point relative to the robot base coordinate system,
    \item $v\_forward$ : the speed at which the robot moves forward,
    \item $v\_turn$ : the speed at which the robot turns,
    \item $eeflx$, $eefly$, $eeflz$ : the position of the robot end effector relative to the robot base coordinate system.
\end{enumerate}
In addition to these environmental variables, the observation space also includes a 220-dimensional LiDAR scan.

\subsection{Pure Manipulation Observation Space}
\label{pure manipulation observation space}
The observation space includes an RGB image with dimensions (64x48x3). In this set of tasks, the observed environmental variables include:
\begin{enumerate}
    \item $pencil\_box\_x$,$pencil\_box\_y$,$pencil\_box\_z$ : the position of the pencil box relative to the robot's base coordinate system,
    \item $eefx$, $eefy$, $eefz$ : the position of the robot end effector relative to the robot base coordinate system,
    \item $target\_eeflz$ : the expected stopping height of the end effector in step 1,
    \item $ori1$, $ori2$, $ori3$, $ori4$ : the current posture of the end effector,
    \item $target\_ori1$, $target\_ori2$, $target\_ori3$, $target\_ori4$ : the target posture of the end effector.
\end{enumerate}

\subsection{Neural Network Architecture}
\label{neural network architecture}
Mobile manipulation task use 1D convolutional neural network to process LiDAR scan data, while pure manipulation task use 2D convolutional neural network to process RGB image data. Below are explanations for some related symbols:
$C1(n, k, s)$, 1D convolution layers, with n being the number of kernels, k being the kernel size, and s being the stride; $C2(n, k, s)$, 2D convolution layers, with n being the number of kernels, k being the kernel size, and s being the stride; $Pool(k, s)$, pool layers, with k being the kernel size, and s being the stride; $F(n)$, fully connected layer;$L$, Flattening.

Both the causal discovery phase and the reinforcement learning training phase require processing of LiDAR data and RGB image data, and the neural network structures used in both phases are the same. The LiDAR scan data is passed through $C1(32, 8, 4)-C1(64, 4, 2)-C1(64, 3, 1)-L$, and the RGB data is passed through $C2(32,(8,6),4)-C2(64,(4,3),2)-C2(32,(8,6),4)-Pool((2,2),2)-C2(64,(2,2),2)-L$. In the causal discovery phase, the output are concatenated with the vector of environmental variables. Then the concatenated vector passes through $F(128)-F(128)-F(128)-(F(1),F(1))$,outputs the mean and logarithm of the standard deviation. In the reinforcement learning training phase, the results of the above feature extraction are also concatenated with other environmental variables. Subsequently, for the mobile manipulation task experiments based on PPO, the data is input into policy networks and value networks with the structure $F(64)-F(64)$. For the pure manipulation task experiments based on SAC, the data is input into policy networks and action-value networks with the same $F(512)-F(512)$.

\subsection{Hyperparameter Setting}
\label{Hyperparameter setting}
In causal discovery phase, learning rate is set to 5e-4, batch size is set to 32. The randomly sampled data is used with 80\% for training the reward prediction model and 20\% for causal discovery. The amount of data collected through intervention sampling is the same as the amount of data collected through random sampling.
For mobile manipulation task, epochs is set to 500. For each action-reward pair, 10,000 samples are randomly collected. For pure manipulation task, epochs is set to 100, 5000 samples are randomly collected. 

The hyperparameters for the reinforcement learning phase are shown in the table \ref{tab:hyperparameter}.
\begin{table}[ht]  
  \caption{Hyperparameter Setting}  
  \renewcommand{\arraystretch}{2}
  \setlength{\tabcolsep}{3pt}
  \begin{center}
  \begin{tabular}{|c|c|c|}  
    \hline  
    \textbf{task} & \textbf{hyperparameter} &\textbf{value}
    \\
    \hline
    \multirow{1}{*}{mobile manipulation}
        & discount factor
        & 0.99
        \\
        \cline{2-3}
        & learning rate
        & 5e-5
        \\
        \cline{2-3}
        & PPO clip range
        & 0.2
        \\
        \cline{2-3}
        & The number of steps sampled in an epoch
        & 10000
        \\
        \cline{2-3}
        & batch size
        & 64
        \\
        \cline{2-3}
        & The number of optimization epochs for each update
        & 10
        \\
        \cline{2-3}
    \cline{1-3}
    \multirow{3}{*}{pure manipulation}
        & discount factor
        & 0.99
        \\
        \cline{2-3}
        &   subtask buffer size
        &   10000
        \\
        \cline{2-3}
        &   The number of training samples for subtasks.
        &   1000
        \\
        \cline{2-3}
        &   batch size
        &   512
        \\
        \cline{2-3}
        &   learning rate
        &   5e-4
        \\
        \cline{2-3}
        &   soft target update factor
        &   0.005
        \\
    \cline{1-3}
  \end{tabular}
  \label{tab:hyperparameter}  
  \end{center}
\end{table}
\end{document}